\documentclass{article}
\usepackage{ijcai24}

\usepackage{times}
\usepackage{soul}
\usepackage{url}
\usepackage[hidelinks]{hyperref}
\usepackage[utf8]{inputenc}
\usepackage[small]{caption}
\usepackage{graphicx}
\usepackage{amsmath}
\usepackage{amsthm}
\usepackage{amssymb}
\usepackage{booktabs}
\usepackage{algorithm}
\usepackage{algorithmic}
\usepackage[switch]{lineno}
\usepackage{color}
\usepackage{natbib}
\usepackage{subcaption}
\usepackage{tikz}
\usepackage{forest}
\usetikzlibrary{trees,positioning,shapes,shadows,arrows.meta}
\usepackage{tcolorbox}
\usepackage{makecell}

\newtheorem*{framework*}{Framework}

\title{Beyond the Speculative Game: \\A Survey of Speculative Execution in Large Language Models}

\author{
    Chen Zhang, Zhuorui Liu, Dawei Song
    \affiliations
    Beijing Institute of Technology, China
    \emails
    \texttt{\{czhang,zrliu,dwsong\}@bit.edu.cn}
}

\begin{document}

\maketitle

\begin{abstract}
% \todo{In addition, at least one author must submit a short CV (up to 2 pages in any reasonable format) providing evidence that they are an expert on the topic of the submitted survey paper. Submitted technical papers must be no longer than nine (9) pages in total: seven pages for the main text of the paper (including all figures but excluding references) and up to two additional pages for references (these may also include acknowledgements, as well as an ethics statement and a contribution statement)} 

With the increasingly giant scales of (causal) large language models (LLMs), the inference efficiency comes as one of the core concerns along the improved performance. In contrast to the memory footprint, the latency bottleneck seems to be of greater importance as there can be billions of requests to a LLM (e.g., GPT-4) per day. The bottleneck is mainly due to the autoregressive innateness of LLMs, where tokens can only be generated sequentially during decoding. To alleviate the bottleneck, the idea of speculative execution, which originates from the field of computer architecture, is introduced to LLM decoding in a \textit{draft-then-verify} style. Under this regime, a sequence of tokens will be drafted in a fast pace by utilizing some heuristics, and then the tokens shall be verified in parallel by the LLM. As the costly sequential inference is parallelized, LLM decoding speed can be significantly boosted. Driven by the success of LLMs in recent couple of years, a growing literature in this direction has emerged. Yet, there lacks a position survey to summarize the current landscape and draw a roadmap for future development of this promising area. To meet this demand, we present the very first survey paper that reviews and unifies literature of speculative execution in LLMs (e.g., blockwise parallel decoding, speculative decoding, etc.) in a comprehensive framework and a systematic taxonomy. Based on the taxonomy, we present a critical review and comparative analysis of the current arts. Finally we highlight various key challenges and future directions to further develop the area.
% , by referring to the available implementations and applications in the open-source community
\end{abstract}

\section{Introduction}

In recent one year or two, (causal) large language models (LLMs) have rushed into people's eyes because of their remarkable performance in terms of reasoning, decision, and so on~\citep{DBLP:conf/nips/BrownMRSKDNSSAA20,DBLP:journals/jmlr/ChowdheryNDBMRBCSGSSTMRBTSPRDHPBAI23,DBLP:journals/corr/TouvronLI23}. However, the benefits come with a critical concern. The inference efficiency, especially the inference latency in comparison to the inference memory~\citep{DBLP:journals/corr/WanWL23}, raises anxieties on whether LLMs could be sustainable with respect to billions of requests per day like those submitted to GPT-4~\citep{DBLP:journals/corr/OpenAI23}. 

It is commonly recognized that the latency is bottlenecked by the autoregressive nature of LLMs, in which tokens are generated in a one-by-one fashion during decoding. To mitigate the bottleneck, a wide range of methods such as pruning~\citep{DBLP:journals/corr/MaFW23,DBLP:journals/corr/XiaGZ23,DBLP:journals/corr/SunLB23}, distillation~\citep{DBLP:journals/corr/TunstallBL23,DBLP:journals/corr/ZhangSY23}, have been proposed. Among these methods, the idea of speculative execution as in computer architecture~\citep{DBLP:books/daglib/HennessyP12} is introduced to LLM decoding in a \textit{draft-then-verify} style. The regime would speculatively obtain a sequence of tokens via some heuristic techniques and then feed the sequence to the LLM so that the tokens could be verified. Unlike sequential generation, the verification of tokens is parallel, which can significantly lift the decoding speed. Distinguished from other methods to circumvent the bottleneck, this is the unique direction that concentrates on the efficiency improvement in the decoding. Due to this, the direction is orthorgnal to any other directions and can be easily combined with others to achieve more significant gains.

While a variety of work has been devoted to optimizing the speculative execution in LLMs (e.g., blockwise parallel decoding that foresees coming tokens in one step with multiple predictive heads~\citep{DBLP:conf/nips/SternSU18}, speculative decoding that formulates a sub-optimal token sequence efficiently with auxiliary small models~\citep{DBLP:conf/icml/LeviathanKM23}, etc.), there is a lack of a systematic critical review of this exciting area. To this end, we present the very first survey that reviews and unifies the literature of LLM speculative execution in a comprehensive framework and a systematic taxonomy. Based on the taxonomy, the existing methods in this area are summarized, critically analyzed and compared, unveiling various key challenges that need to be tackled in the future. %Through discussing widespread implementations and applications in the open-source community, 
Correspondingly, we highlight a series of promising directions and opportunities that can be explored to further develop this encouraging yet challenging area.\footnote{We note that there is a concurrent survey of speculative decoding released right before (Jan. 15th, 2024) our submission at \texttt{\url{https://arxiv.org/abs/2401.07851}}, However, it takes a much looser framework than ours.}

\begin{figure*}[ht]
    \centering
    \begin{subfigure}[]{0.43\textwidth}
         \centering
         \includegraphics[width=0.77\textwidth]{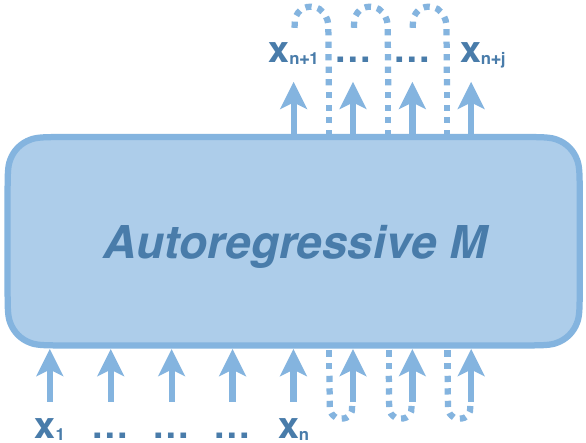}
         \caption{Autoregressive decoding, where tokens are sequentially decoded.}
         \label{fig:spec_autoregressive}
    \end{subfigure}
    \qquad
    % \hfill
    \begin{subfigure}[]{0.43\textwidth}
         \centering
         \includegraphics[width=0.77\textwidth]{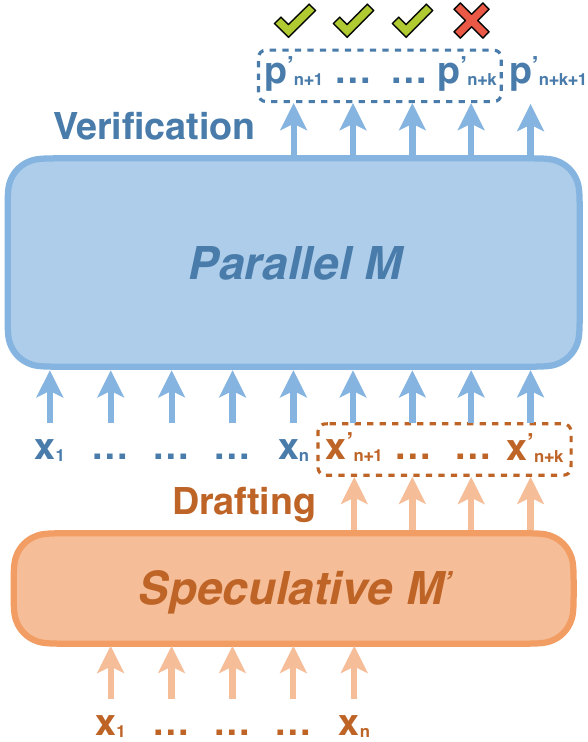}
         \caption{Speculative execution, where tokens are speculatively drafted and in parallel verified.}
         \label{fig:spec_speculative}
    \end{subfigure}
    \caption{An illustration of speculative execution in LLMs via a comparison between autoregressive decoding and speculative execution.}
    \label{fig:spec_exec}
\end{figure*}

The survey is organized as follows: first, we offer an overview of LLM decoding and speculative execution in LLMs (Section~\ref{sec:spec_exec}), which can be decomposed into the designs of drafting and verification; second, we  delve into the design of drafting (Section~\ref{sec:drafting} and design of verification (Section~\ref{sec:verification}) respectively; then, we showcase how to evaluate the LLM speculative execution systems (Section~\ref{sec:eval}) and discuss some well-known implementations and applications (Section~\ref{sec:impl_appl}). Finally, we conclude the key challenges and opportunities of speculative execution in LLMs (Section~\ref{sec:chal_oppo}).

\section{Speculative Execution in LLMs}
\label{sec:spec_exec}

A critical issue that the causal LLMs are faced with is the latency problem. Owing to the autoregressive nature of LLMs, their decoding can suffer from pretty high latency, making it resource-intensive to handle billions of requests per day. Formally, given a sequence of input tokens $I=\{x_i\}_{i=1}^{n}$ and a target model $\mathcal{M}$, the autoregressive decoding process can be described as below:
\begin{equation}
    x_{n+j}\sim\mathcal{M}(x|x_{1},\dots,x_{n},x_{n+1},\dots,x_{n+j-1})
\end{equation}
where $x_{n+j}$ is the $j$-th output token to be decoded. It is the dependency of $x_{n+j}$ to all previous input and already generated tokens that makes the LLM decoding  autoregressive, as shown in Figure~\ref{fig:spec_autoregressive}.

Speculative execution is a concept originally proposed in the field of computer architecture~\citep{DBLP:books/daglib/HennessyP12}. It is a strategy that fully utilizes spare resources to execute some speculative tasks in advance that may or may not be useful for the upcoming tasks. If the speculatively executed tasks are useful, the execution speed of upcoming tasks would be substantially enhanced. Otherwise, the speed would not be degraded but remain the same as the circumstance that these speculative tasks have never been executed. Inspired by this idea, speculative execution is adapted to LLMs in expectation to speed-up LLM decoding.

The adaptation primarily follows a \textit{draft-then-verify} paradigm. The drafting stage is akin to executing the speculative tasks, and the verification stage is similar to checking the usefulness of the speculative tasks in the upcoming tasks. In the language of LLM decoding, the drafting stage accords with a fast construction of a sequence of tokens, and the verification stage corresponds to a parallel validation of the sequence of tokens. Since there is no obvious spare time in LLM decoding, the drafting stage should be as fast as possible. If the tokens from the drafting stage are of high quality, then the verification stage would benefit from the drafting and be accelerated by the virtue of shifting from the autoregressive decoding to the parallel verification. By notations, the drafting stage can be interpreted as below:
\begin{equation}
    x_{n+j}^{\prime}\sim\mathcal{M}^{\prime}(x|x_{1},\dots,x_{n},x_{n+1}^{\prime},\dots,x_{n+j-1}^{\prime})
\end{equation}
where $\mathcal{M}^{\prime}$ is a fast and speculative drafter. There could be multiple drafts rather than only one draft for an improved performance, and these drafts will be appropriately managed. After that, the managed speculative tokens $T=\{x_{n+j}^{\prime}\}_{j=1}^{k}$ will be verified by the target model (or say verifier) $\mathcal{M}$ to determine whether the speculative tokens are useful or not as below:
\begin{equation}
    P=\{p_{k+1}^{\prime}\}_{j=1}^{k+1}\sim\mathcal{M}(x_{1},\dots,x_{n},x_{n+1}^{\prime},\dots,x_{n+k}^{\prime})
\end{equation}
where $P$ is the speculative probabilities based on the speculative tokens. Then an acceptance criterion will be applied by taking both $T$ and $P$ into considerations to determine whether the speculative tokens are valid or not and yield the final output $O$. Typically, only part of the speculative tokens are valid. The partly-accepted case is also viewed as a useful speculation. In view of this, a too long or too short speculation may be less useful, as a short speculative output may degenerate to autoregressive decoding (e.g., $k$ is 1) and a long speculative output may give diminishing speed-up return yet heavy latency cost. Therefore, commonly a termination criterion will be deployed in the drafting stage for an early stop of the drafter.

\begin{figure*}
    \centering
    \tikzset{
        basic/.style  = {draw, text width=3cm, align=center, font=\sffamily, rectangle},
        root/.style   = {basic, rounded corners=2pt, thin, align=center, fill=green!20, text width=3cm,},
        xnode/.style = {basic, thin, rounded corners=2pt, align=center, fill=cyan!20, text width=2cm,},
        tnode/.style = {basic, thin, rounded corners=2pt, align=left, fill=pink!30, text width=2cm, align=center},
        onode/.style = {basic, thin, rounded corners=2pt, align=center, fill=orange!30,text width=3.7cm,},
        edge from parent/.style={draw=black, edge from parent fork right}
    }
    \begin{forest} for tree={
        grow=east,
        growth parent anchor=west,
        parent anchor=east,
        child anchor=west,
        edge path={\noexpand\path[\forestoption{edge},->, >={latex}] 
             (!u.parent anchor) -- +(10pt,0pt) |-  (.child anchor) 
             \forestoption{edge label};}
    }
    % l sep is used for arrow distance.
    [Speculative Execution in LLMs, root,  l sep=7mm,
        [Design of Verification, xnode,  l sep=7mm,
            [Acceptace Criterion, tnode,  l sep=7mm,
                [Typical Acceptance, onode]
                [Rejection Sampling, onode]
                [Exact Matching, onode]
            ]
            [Verifier, tnode,  l sep=7mm,
                [Tree, onode]
                [Chain, onode]
            ]
        ]
        [Design of Drafting, xnode,  l sep=7mm,
            [Draft \\Management, tnode,  l sep=7mm,
                [Trie Tree, onode]
                [Beam Tree, onode]
                [Linear Chain, onode]
            ]
            [Termination Criterion, tnode,  l sep=7mm,
                [Heuristic Rule, onode]
                [Adaptive Thresholding, onode]
                [Static Setting, onode ]
            ]
            [Drafter, tnode,  l sep=7mm,
                [Retrieval, onode]
                [Generation, onode]
            ]
        ]
    ]
    \end{forest}
    \caption{A systematic taxonomy of existing literature of speculative execution in LLMs.}
    \label{fig:taxo}
\end{figure*}
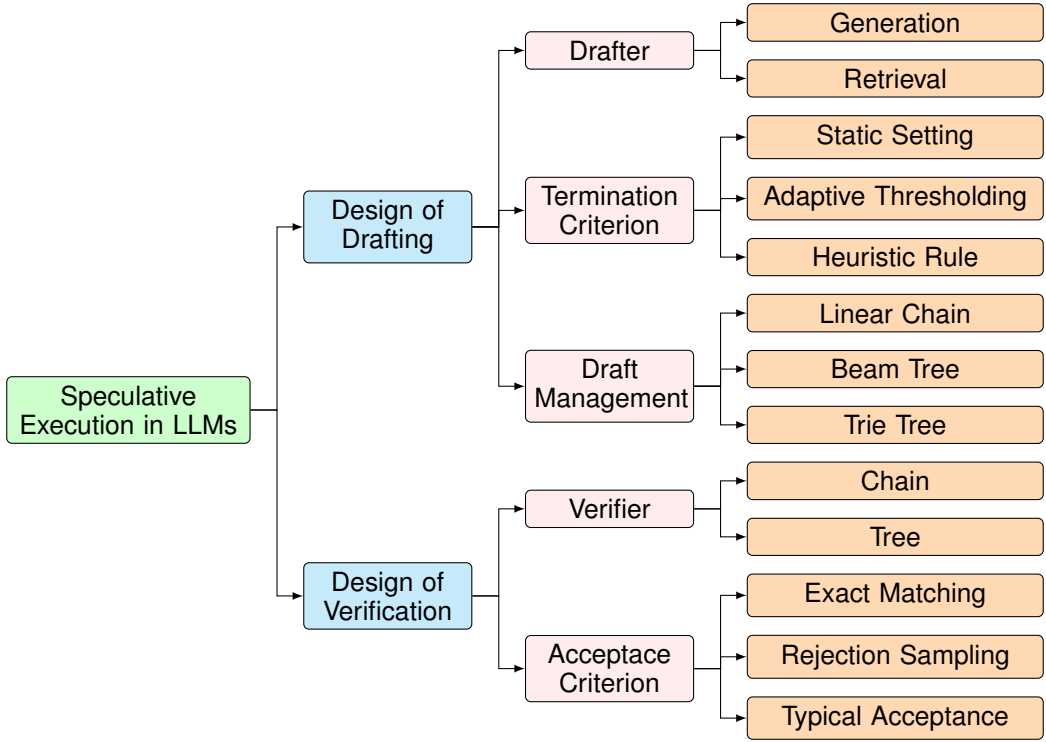

The above description provides an overview on speculative execution in LLMs, as shown in Figure~\ref{fig:spec_speculative}. Deviated from the overview, we can draw a comprehensive framework, as formulated below:
\begin{tcolorbox}[width=0.477\textwidth,colback={green!20}] 
\begin{framework*}
Given an input $I$, the drafter $\mathcal{M}^{\prime}$ drafts one or multiple sequences of managed tokens $T$ in a speculative manner in the drafting stage. Then the verifier based on this further produces a sequence of probabilities $P$, verifies the usefulness of the speculative tokens, and leads to the final output tokens $O$ in parallel. Proper termination criterion should be settled for the drafter, as is the acceptance criterion for the verifier. Conceptually,
\begin{equation}\nonumber
\begin{aligned}
    &\textbf{Drafting: }I\rightarrow\mathcal{M}^{\prime}\xrightarrow{\text{termination}}T \\
    &\textbf{Verification: }I\circ T\rightarrow\mathcal{M}\rightarrow P\xrightarrow{\text{acceptance}}O
\end{aligned}
\end{equation}
where $\circ$ denotes a combination of the input $I$ and the speculative tokens $T$.
\end{framework*}
\end{tcolorbox}

While each involved component in the framework can vary from one to another, the variations can be generally categorized according to the concerned stages. This grants us a chance of setting up a systematic taxonomy of existing literature, as shown in Figure~\ref{fig:taxo}. Following this taxonomy, we can organize the existing methods into different categories, which will be detailed in the next sections.

\section{Design of Drafting}
\label{sec:drafting}
% \todo{seemingly lacking of critiques.}

This category mainly involves the design of the drafter, termination criterion, and draft management. The drafter is abstracted as a draft constructer, which serves as the core of the drafting stage.  The termination criterion is an indicator of when the drafter should be terminated, or in other words, how the length of the speculative tokens $k$ is chosen. In view that there might be multiple drafts instead of a unique one, the draft management is aimed at organizing these drafts.

\subsection{Drafter}

\paragraph{Generation versus Retrieval}

\begin{figure*}[ht]
    \centering
    \begin{subfigure}[]{0.3\textwidth}
        \centering
        \includegraphics[width=0.97\textwidth]{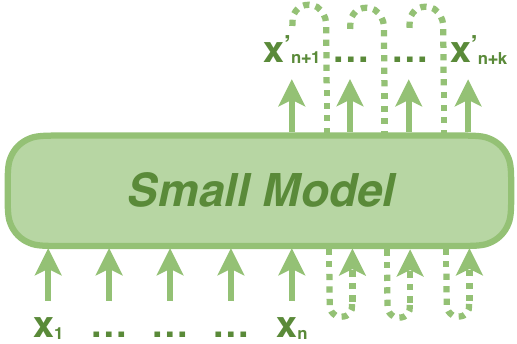}
        \caption{Speculative execution with a small model, alias speculative decoding.}
        \label{fig:drafter_small_model}
    \end{subfigure}
    \quad
    \begin{subfigure}[]{0.3\textwidth}
        \centering
        \includegraphics[width=0.97\textwidth]{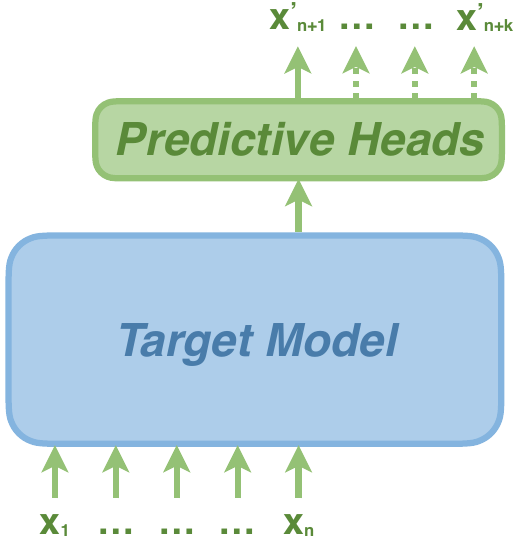}
        \caption{Speculative execution with multiple predictive heads, alias blockwise parallel decoding.}
        \label{fig:drafter_predictive_head}
    \end{subfigure}
    \quad
    \begin{subfigure}[]{0.33\textwidth}
        \centering
        \includegraphics[width=0.99\textwidth]{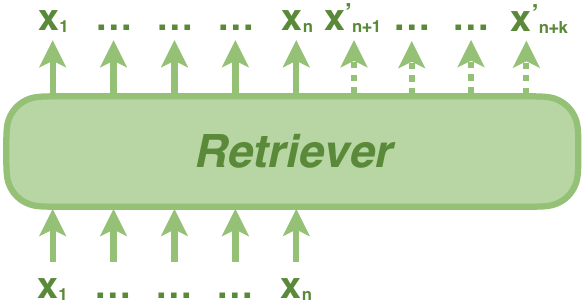}
        \caption{Speculative execution with a retriever, alias retrieval-based speculative decoding.}
        \label{fig:drafter_retriever}
    \end{subfigure}
    \caption{The illustration of different categories of the drafter.}
    \label{fig:drafter}
\end{figure*}

To certain degree, the drafter itself is usually a causal language model that generates speculative tokens. The drafter can be an additional small model besides the target model that generates candidate tokens as in speculative decoding~\citep{DBLP:conf/icml/LeviathanKM23} or several lightweight predictive heads attached to the target model that foresee the forthcoming (e.g., next-next) tokens as in blockwise parallel decoding~\citep{DBLP:conf/nips/SternSU18}. Recent progresses suggest that the drafter can also be a retriever that retrieves tokens from a large corpus that complete the preceding context~\citep{DBLP:journals/corr/HeZC23}, as indicated in Figure~\ref{fig:drafter_small_model} versus~\ref{fig:drafter_retriever}. The retriever is not only faster than the small model and the predictive heads, but also promising if the corpus is of high merit (e.g., high quality and relevance). 

\paragraph{Small Models versus Predictive Heads}

As mentioned above, the central difference between blockwise parallel decoding and speculative decoding lies in their model usages, as indicated in Figure~\ref{fig:drafter_small_model} versus~\ref{fig:drafter_predictive_head}.

The speculative decoding requires additional small models that autoregressively draft speculative tokens. These small models are constrained and are thus much more efficient than the target model, so that the speed-ups can cover the costs of them.

The construction of the small models can be based on existing models or derived models. It should be highlighted that the small models must have exactly the same vocabularies as the target model, so there could be few existing models that can be directly used in the drafting~\citep{DBLP:journals/corr/ChenBI23}. This naturally entails the derivation of the small models from the target model via distillation~\citep{DBLP:journals/corr/ZhouLR23}, quantization~\citep{DBLP:journals/corr/MiaoOZ23}, layer-skipping~\citep{DBLP:journals/corr/ZhangWL23}, and early-exiting~\citep{DBLP:journals/corr/HopperKM23}, etc.

Moreover, the drafting is not limited to only one small model. It is argued that staged or cascaded small models of different scales can further promote the performance, motivated by ensemble learning~\citep{DBLP:journals/corr/SpectorR23,DBLP:journals/corr/ChenYL23}. Staged small models~\citep{DBLP:journals/corr/SpectorR23} somehow introduce a pre-verification process where the speculative tokens proposed by a smaller model is verified by a slightly larger (but still small compared with the target LLM) model. This process will iterate until all small models are staged through and  will produce speculative tokens of higher quality with minor time overhead. On the other hand, cascaded small models~\citep{DBLP:journals/corr/ChenYL23} posit that the earlier speculative tokens should be harder to draft than the later ones, and thus allocates slightly larger models for the earlier tokens yet smaller models for the later tokens.

The blockwise parallel decoding urges lightweight predictive heads to draft speculative tokens in a prophetic manner. While the prediction needs an execution of the target model, multiple tokens instead of one token are produced and the drafting process can be potentially combined with the verification process in one run (i.e., drafting during verification is possible). 

It is evident that simple duplication and tuning of the original prediction head for language modeling should be already very effective~\citep{CaiLG23}. However, latest advances have claimed that more complex head architectures~\citep{LiZZ23} are more appealing. Moreover, using only one predictive head that conditions on different looking-ahead tokens to mimic multiple predictive heads is also promising~\citep{DBLP:journals/corr/MoneaJG23}.

\subsection{Termination Criterion}

\paragraph{Static Setting}

As mentioned, either a too short or too long sequence of speculative tokens is sub-optimal. Without specifications on the length, the drafter is not properly regulated. A straightforward solution to this issue is to set the length $k$ to a static value that may be iteratively and manually refined~\citep{DBLP:journals/corr/HeZC23}.

\paragraph{Adaptive Thresholding}

Although static setting can fulfill most use cases, the labor intensity brought by it could also be cumbersome. To address the problem, adaptive thresholding methods have been proposed, aiming to early stop the drafting based on the per-token confidence~\citep{DBLP:journals/corr/KimMM23}. If the confidence is below a threshold, the drafting will be stopped. The threshold can be adaptively adjusted with respect to some optimization objective (e.g., the drafting quality)~\citep{DBLP:journals/corr/ZhangWL23}.

\paragraph{Heuristic Rules}

Furthermore, a few heuristic rules can also be engaged in the termination. A typical design is that the length of the speculative tokens will be increased if the previous speculation is fully accepted in the verification, and otherwise it will be decreased~\citep{Gante23}. Another approach could be altering the length depending on the batch size from the system serving perspective~\citep{DBLP:journals/corr/SuGP23}.

\subsection{Draft Management}

\paragraph{One Draft versus Multiple Drafts}

It is identified that bringing only one draft during drafting can inhibit the usefulness of the speculative execution. Thus the use of multiple drafts is also an interesting topic. Consequently, draft management serves as a bridge to organize these drafts and export drafts to the verification stage.

There are many ways to generate multiple drafts. For retrieval-based drafters, it is  spontaneous to get multiple drafts since the retrieval could always return multiple candidates. For generation-based drafters, the most naive way to come up with multiple drafts is sampling multiple times~\citep{DBLP:journals/corr/SunSR23}. Another approach  is using multiple heterogeneous small models or predictive heads if available~\citep{DBLP:journals/corr/MiaoOZ23}. Currently, there is also an outgrowth trend to leverage a beam of top-k tokens to form a tree of candidates~\citep{CaiLG23,YangHD24}. 

Necessarily, multiple drafts would demand corresponding tactics in the verification stage to import the generated drafts properly, which will be detailed in Section~\ref{sec:verifier}.

\section{Design of Verification}
\label{sec:verification}

This category involves  the design of a verifier and an acceptance criterion. The verifier is usually the target model with specifications according to the structure of potentially multiple drafts. The acceptance criterion aims to judge whether or not the drafts should be partly accepted, i.e., whether it is fine that the length of the accepted tokens is smaller than $k$.

\subsection{Verifier}
\label{sec:verifier}

\paragraph{Chain versus Tree}

\begin{figure}[ht]
    \centering
    \includegraphics[width=0.39\textwidth]{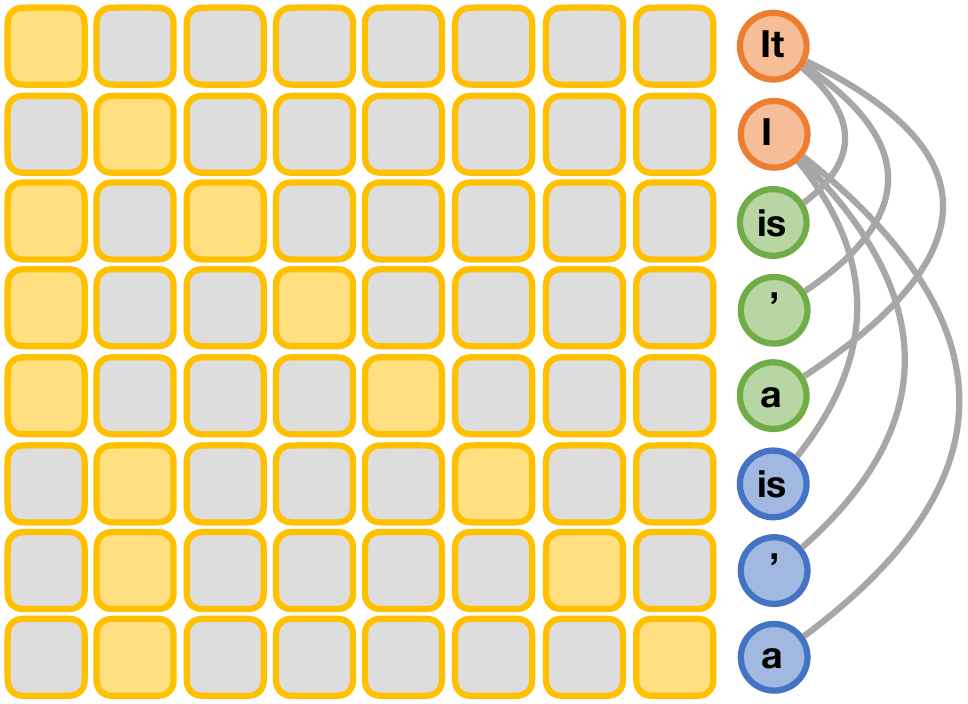}
    \caption{The illustration of tree attention. In the tree attention, child tokens can only see their parent tokens, facilitating the parallel verification of potentially multiple drafts. In contrast, one draft would only require a chain attention, which is causal and lower-triangular.}
    \label{fig:verifier_tree}
\end{figure}

If only one draft needs to be verified, then a chain-based verifier, which essentially means a common one that takes in tokens as a sequence or a chain, should be fairly enough.  Nonetheless if multiple drafts are used, consecutively verifying these drafts one by one would be too time-consuming. Thereby, a tree-based verifier is imposed and this idea is largely shared across literature~\citep{DBLP:journals/corr/MiaoOZ23,DBLP:journals/corr/HeZC23,CaiLG23}. For example,~\citet{DBLP:journals/corr/HeZC23} establishes a trie tree from multiple drafts, prunes less frequent nodes from the trie tree, and verifies them at one run with a tree attention (i.e., the child tokens can only see their parent tokens by attention masking), as presented in Figure~\ref{fig:verifier_tree},

\subsection{Acceptance Criterion}

\paragraph{Exact Matching}

Once a draft is fed into the target model, we can attain the corresponding output probabilities. By aligning the speculative tokens and the probabilities, we can infer whether each token is valid or not in the draft. The most simple acceptance criterion is exact matching~\citep{Gante23}, which examines whether a speculative token is correspondingly the one with the maximum probability.

\paragraph{Rejection Sampling}

Although exact matching can guarantee that the verified output shares the same quality with the output of the target model itself at a minor cost, this equality holds only when greedy decoding is used. For cases where sampling decoding is used for the target models, exact matching can hardly accept tokens from the draft, which may result in a slow-down rather than a speed-up of decoding. Hence, an acceptance criterion modified from rejection sampling is proposed to alleviate the issue~\citep{DBLP:conf/icml/LeviathanKM23,DBLP:journals/corr/ChenBI23}. In theory, this acceptance criterion can be applied to both greedy decoding and sampling decoding. Yet, if multiple drafts are given, this acceptance criterion can unexpectedly accept a few of them so that we cannot trivially determine which one to use actually. So an optimal transport perspective is studied towards improving the criterion~\citep{DBLP:journals/corr/SunSR23}.

\paragraph{Typical Acceptance}

The above-mentioned two acceptance criteria provide rigorous guarantees for quality. However, a too rigorous acceptance criterion can offset the effort of the parallel verification and degrade the speculative execution as a burden, especially when a temperature is applied. Consequently, a moderately relaxed acceptance criterion is needed for more observable speed-up in some scenarios. Typical acceptance takes this role and accepts tokens in the draft if their speculative probabilities go beyond a hard threshold~\citep{CaiLG23}. In another perspective~\citep{DBLP:conf/emnlp/Xia0WCWS23}, the threshold is dynamically adjusted via a top-k constraint. For the case that multiple drafts are offered, typical acceptance will consider the one that forms the longest sequence and drop others. Critically, typical acceptance can lead to unfair comparisons in experiments and should be properly highlighted if used. 

\section{Evaluation}
\label{sec:eval}

\begin{table*}[ht]
    \centering
    \caption{The reported speed-ups in comparison with autoregressive decoding from available representative methods. Tran stands for translation task, Summ stands for summarization task, and Inst stands for instruction-following task.}
    \label{tab:perf}
    \begin{tabular}{lrrrrrr}
    \toprule
        \textbf{Method} & \textbf{Drafter} & \textbf{Verifier} & \textbf{Task} & \textbf{Target Model} & \textbf{Speed-up} \\
    \midrule
        Vicuna~\citeyearpar{ChiangLL23} & -- & -- & -- & -- & 1.0$\times$ \\
        Blockwise Decoding~\citeyearpar{DBLP:conf/nips/SternSU18} & predictive heads & chain & Tran & Transformer & $\sim$3.3$\times$ \\
        SpecInfer~\citeyearpar{DBLP:journals/corr/MiaoOZ23} & multiple small models & tree & Inst & LLaMA & $\sim$2.0$\times$  \\
        Medusa~\citeyearpar{CaiLG23} & predictive heads & tree & Inst & Vicuna & $\sim$1.9$\times$ \\
        Eagle~\citeyearpar{LiZZ23} & predictive heads & tree & Inst & Vicuna & $\sim$3.0$\times$ \\
        SpecDec~\citeyearpar{DBLP:conf/emnlp/Xia0WCWS23} & look-ahead tokens & chain & Tran & Transformer & $\sim$3.0$\times$ \\
        BiLD~\citeyearpar{DBLP:journals/corr/KimMM23} & small model & chain & Tran/Summ & mT5 & $\sim$1.6$\times$ \\
        Speculative Decoding~\citeyearpar{DBLP:conf/icml/LeviathanKM23} & small model & chain & Tran/Summ & T5 & $\sim$2.5$\times$\\
        Speculative Sampling~\citeyearpar{DBLP:journals/corr/ChenBI23} & small model & chain & Summ/Code & Chinchilla & $\sim$2.2$\times$ \\
        Assisted Generation~\citeyearpar{Gante23} & small model & chain & Summ & OPT & $\sim$2.6$\times$ \\
        Staged Speculative~\citeyearpar{DBLP:journals/corr/SpectorR23} & staged small models & chain & Code & GPT-2 & $\sim$1.6$\times$ \\
        SpecTr~\citeyearpar{DBLP:journals/corr/SunSR23} & small model & tree & Inst & PaLM & $\sim$2.1$\times$ \\
        
        Self-Speculative~\citeyearpar{DBLP:journals/corr/ZhangWL23} & layer-skipped small model & chain & Summ/Code & LLaMA-2 & $\sim$1.4$\times$ \\
        DistilSpec~\citeyearpar{DBLP:journals/corr/ZhouLR23} & distilled small model & chain & Inst/Summ/Math & GPT-2/T5 & $\sim$1.3$\times$ \\
        REST~\citeyearpar{DBLP:journals/corr/HeZC23} & retrieval & tree & Code/Inst & LLaMA/Vicuna & $\sim$1.9$\times$ \\
        Cascaded Speculative~\citeyearpar{DBLP:journals/corr/ChenYL23} & cascaded small models & chain & Inst/Math & T5 & $\sim$3.5$\times$ \\
        MultiCand Speculative~\citeyearpar{YangHD24} & small model & tree & Inst/Tran & LLaMA/Vicuna & $\sim$2.2$\times$ \\
        PaSS~\citeyearpar{DBLP:journals/corr/MoneaJG23} & look-ahead tokens & chain & Code & LLaMA & $\sim$1.5$\times$ \\
    \bottomrule
    \end{tabular}
\end{table*}

To evaluate the performance of speculative execution in LLMs, a common evaluation measure is the latency per token, which is computed as the elapsed time divided by the number of output tokens. It is formally defined as below:
\begin{equation}
    \text{latency}=\frac{\text{elapsed seconds}}{\text{\#generated tokens}}
\end{equation}
In some literature, a variant of this measure is used in its reciprocal form, i.e., tokens per second as below:
\begin{equation}
    \text{throughput}=\frac{\text{\#generated tokens}}{\text{elapsed seconds}}
\end{equation}
After obtaining these measures, the relative speed-up can be viewed as an important comparative measurement between two methods.

Commonly used datasets for evaluation range from instruction-following ones like Alpaca~\citep{TaoriGZ23} and MT-Bench~\citep{DBLP:journals/corr/ZhengCS23}, to mathematics data such as  GSM8K~\citep{DBLP:journals/corr/CobbeKB21}, and coding data like HumanEval~\citep{DBLP:journals/corr/ChenTJ21}. Others may include summarization datasets like CNN/DailyMail~\citep{DBLP:conf/acl/SeeLM17}, XSum~\citep{DBLP:conf/emnlp/NarayanCL18} and translation datasets like WMT.

Hereby, we offer an overview of reported performance of the methods that are reviewed in this paper, as a cheatsheet displayed in Table~\ref{tab:perf}.

\section{Implementations and Applications}
\label{sec:impl_appl}

There are various implementations of speculative execution in a range of LLM inference frameworks. Likewise, the applications are also not unique. %We will inspect their practical utilization.

\subsection{Implementations}

\paragraph{LLaMA.cpp}

To facilitate the widespread accessibility of Large Language Models (LLMs), we introduce LLaMA.cpp~\citep{llamacpp23}, a robust solution designed to streamline the deployment process. LLaMA.cpp enables the users to efficiently operate an LLM across various platforms, leveraging advanced model compression techniques like quantization. Specifically, LLaMA.cpp features a comprehensive integration of C/C++ implementations devoid of external dependencies. It employs a mixed-precision computation approach, and offers a range of quantization options. Notably, LLaMA.cpp has been adeptly adapted for deployment on Apple silicon. This innovation allows users, even those with limited GPU resources, to run an LLM on their personal computers effectively.

In parallel, speculative execution serves as a straightforward yet effective method to reduce inference latency. This library includes various implementations of speculative execution, such as standard speculative decoding and tree-based speculative decoding. We will elucidate these methods in a step-by-step manner. Fundamentally, both approaches adhere to a similar process, encompassing a drafting stage and a verification stage. During the drafting phase, each utilizes a smaller model to generate speculative tokens. The termination criterion is based on adaptive thresholding, i.e., if the probability of the current token falls below that of the subsequent token or an end-of-sentence token is reached. The primary distinction between these methods lies in their draft management and verification mechanisms in the subsequent stage: the tree-based approach employs tree draft management and a tree verifier, whereas the standard method utilizes a linear chain for both management and verification. In terms of acceptance criteria, an exact matching strategy is implemented for both methodologies.

\paragraph{vLLM}

There have been significant efforts to optimize LLM computations beyond traditional deployment methods because of high latency and substantial memory footprint of LLMs. A notable contribution in this area is vLLM~\citep{kwon2023efficient}, a framework designed for fast and cost-effective inference, addressing challenges related to memory utilization and I/O communication bottlenecks. Central to vLLM is the innovative \textit{Paged Attention} mechanism, which optimizes KV-cache memory use by transforming continuous memory requirements into a discontinuous format, akin to the paged memory systems in operating systems, thereby enhancing overall memory efficiency. A key feature of Paged Attention is its facilitation of efficient memory sharing for parallel sampling, allowing different sequences to map their logical blocks onto the same physical block. In practical deployments, vLLM has demonstrated a remarkable capability, with backend serving throughput reaching up to 30 times higher than that of the traditional HuggingFace backend. 

Within the vLLM's speculative execution framework, various strategies are employed at different stages. In the drafting stage, a smaller model is utilized to generate candidate tokens, followed by the application of a static setting to terminate generation. For draft management, a linear chain strategy has been adopted. The verification stage features the use of a chain scheme and rejection sampling method, consistent with the methodologies presented in original research papers \citep{DBLP:conf/icml/LeviathanKM23, DBLP:journals/corr/ChenBI23}. Importantly, these implementations represent just a preliminary foray into the speculative execution capabilities of vLLM, with ongoing development and additional proposals (PRs) for further enhancements in speculative execution within the vLLM framework. LLaMA.cpp is primarily focused on enhancing the accessibility of LLMs, whereas vLLM concentrates on increasing the throughput and reducing the latency of LLMs deployments. With specific regard to speculative execution, LLaMA.cpp offers a more comprehensive implementation compared to that of vLLM at present.

\subsection{Applications}

\paragraph{Online Speculative Decoding}

It has been observed that constant speculative decoding cannot meet the requirement of continuously changing environment in online services, e.g., queries from users can have diverse topics. This distributional shift at time span can prevent speculative decoding achieving speed-up, and potentially make speculative decoding a burden for serving.

Therefore,~\citet{DBLP:journals/corr/LiuHB23} proposes to continually updating the small model in the speculative decoding with online data. In doing so, the small model is kept updated with the requirement of the online environment.

\paragraph{Contrastive Speculative Decoding}

Contrastive decoding~\citep{DBLP:conf/acl/LiHFLEHZL23} is a technique imposed to boost the generation quality by leveraging the generation probability of a small model as the baseline and contrasting the generation probability of the target model to the baseline. Since problems like hallucination may underlie the probability of the small model, the contrast could result in a better generation quality. 

Naturally, a combination of contrastive decoding and speculative decoding can be used, as both of them utilize an additional small model besides the target model as in~\citet{DBLP:journals/corr/YuanLH23}. It is reported that by incorporating the small model to the target model, both the generation quality and speed are lifted up by a multiple of around 2.

\paragraph{On-device Speculative Decoding}

Owning to the speed-ups brought by speculative execution in LLMs, it is not surprising that speculative decoding is integrated to on-device scenarios to better play with the limited resources~\citep{DBLP:journals/corr/XuYJ23}.
By doing so, LLMs could be executed much faster than ever on mobile devices.

\section{Challenges and Opportunities}
\label{sec:chal_oppo}

In the above sections we have reviewed the state-of-the-art methods and compared the similarities and dissimilarites among them. Based on the observations and discussions, we can now summarize the should-be-highlighted challenges yet to be tackled in the future, whilst the opportunities of exploring possible solutions to these challenges:

\paragraph{CP 1: Framework Design}

Given so many design components to be determined in a speculative execution algorithm, it turns out challenging to customize an optimal framework for speculative execution in a wide range of application scenarios. For example, whether we should use one small model or multiple predictive heads in the drafting is largely variant from one case to another. However, this also offers the opportunity to further optimize and unify currently used frameworks so that the number of such variables would be reduced. This shall spare a lot efforts in finding a design choice for a different task.

\paragraph{CP 2: Parameter Search}

It is identified that there are quite a few parameters to be settled, like the number of speculative tokens and the threshold in typical acceptance. Although various methods have been developed to automatically detect the ideal values of these parameters, it is still hard to tell whether they are good enough. In this demand, we would better  establish more robust methods to search and set these parameters so that more stable and appealing performance can be yielded.

\paragraph{CP 3: System Integration}

While we have analyzed several systems that have integrated the speculative execution for LLMs, more systems are coming out with diverse and advanced properties that have not been considered in the design of speculative execution yet. Therefore, taking into consideration the highlighted features in these new systems will be a never-ending track that should be noted by the research community.

\paragraph{CP 4: Objective Optimization}

Currently, speculative execution in LLMs are mainly utilized to enhance the open-domain generation speed. Nonetheless, different settings would possibly require different attention mechanisms to the pick of objectives. For example, the combination of speculative decoding and contrastive decoding would benefit overcoming the hallucination problem with reduced generation latency. In view of this, flexible combinations to or proper ablations from speculative execution in LLMs should be raised if necessary for specific objectives.

Finally, we would like to conclude that speculative execution is an important and promising area where a range of challenges are yet to be tackled.

% \appendix

% \section*{Ethical Statement}

% There are no ethical issues.

% \section*{Acknowledgments}

\bibliographystyle{named}
\bibliography{ijcai24}

\end{document}